\documentclass{article}

\usepackage{iclr2026_conference,times}
\usepackage{amsmath,amsfonts,bm}

\def\eqref#1{equation~\ref{#1}}
\def\1{\bm{1}}

\DeclareMathAlphabet{\mathsfit}{\encodingdefault}{\sfdefault}{m}{sl}
\SetMathAlphabet{\mathsfit}{bold}{\encodingdefault}{\sfdefault}{bx}{n}

\newcommand{\E}{\mathbb{E}}

\newcommand{\R}{\mathbb{R}}

\usepackage[utf8]{inputenc}
\usepackage[T1]{fontenc}
\usepackage{hyperref}
\usepackage{url}
\usepackage{booktabs}
\usepackage{amsmath,amssymb,amsthm}
\usepackage{mathtools}
\usepackage{bm}
\usepackage{array}
\usepackage{cleveref}
\usepackage{microtype}
\usepackage{xcolor}
\usepackage{multirow}
\usepackage{algorithm}
\usepackage{algorithmic}
\usepackage{graphicx}
\usepackage{subcaption}
\usepackage{float}
\newcommand{\cD}{\mathcal{D}}

\newcommand{\cM}{\mathcal{M}}

\newcommand{\cO}{\mathcal{O}}

\newcommand{\cL}{\mathcal{L}}

\newcommand{\fwrite}{f_{\mathrm{write}}}
\newcommand{\fread}{f_{\mathrm{read}}}
\newcommand{\thetaw}{\theta_{\mathrm{w}}}
\newcommand{\thetar}{\theta_{\mathrm{r}}}

\definecolor{romemTeal}{HTML}{2F7774}
\definecolor{romemSlate}{HTML}{5F6C6B}
\definecolor{romemSand}{HTML}{D8A45E}
\definecolor{romemBoxBg}{HTML}{F4F7F6}
\definecolor{romemBoxLine}{HTML}{AEB9B6}
\newcommand{\romemqualbox}[2]{%
\begingroup
\setlength{\fboxsep}{5pt}%
\noindent\fcolorbox{romemBoxLine}{romemBoxBg}{%
\begin{minipage}{0.91\columnwidth}
\small
\textbf{\textcolor{romemTeal}{#1}}\par\vspace{2pt}
#2
\end{minipage}}%
\endgroup}

\title{Rosetta Memory: Adaptive Memory for \\ Cross-LLM Agents}

\author{\textbf{Hao Yang$^{1}$}, \textbf{Shiqi Shen$^{2}$}, \textbf{Haoxuan Li$^{3}$}, \textbf{Zhipeng Wang$^{2}$}, \textbf{Zhi Gong$^{2}$}, \textbf{Xu Chen$^{1,*}$}\\
$^{1}$Gaoling School of Artificial Intelligence, Renmin University of China, \\
$^{2}$Weixin, Tencent,
$^{3}$Institute for Artificial Intelligence, Peking University\\
\texttt{hao.yang@ruc.edu.cn, shiqishen@tencent.com}\\
\texttt{markrocwang@tencent.com, davidgong@tencent.com}\\
\texttt{hxli@stu.pku.edu.cn, xu.chen@ruc.edu.cn}
}

\iclrfinalcopy
\begin{document}
\maketitle
\begingroup
\renewcommand{\thefootnote}{\fnsymbol{footnote}}%
\footnotetext[1]{Corresponding author.}%
\endgroup
\begin{abstract}
Memory is the key component for transforming a stateless LLM into a persistent, evolving agent through experience accumulation, long-horizon planning, and continual self-improvement.
Existing memory systems typically take the LLM as the center and design memory operations tailored to a specific backbone.
In practice, however, users frequently switch between LLMs, for example using Claude for coding and GPT for writing across tasks, or routing different steps to different backbones within a single task for cost-effective trade-offs.
As a result, memory written by one model often needs to be consumed by another.
Making upstream memory effectively adapt to and activate downstream LLMs remains a critical yet underexplored problem.
To bridge this gap, we shift the perspective from LLM-centric memory design to \emph{memory-centric LLM adaptation}.
Specifically, we approach the above upstream-downstream memory adaptation problem from both the write and read sides, and design two profile-conditioned operators that are jointly trained to optimize how memory is stored and presented for better task completion.
To ensure the learned operators generalize across a broad set of LLMs, we propose a minimum-gain sampling curriculum that prioritizes the least-served LLMs during training.
To better measure the operators' actual contribution rather than the LLM's own capability, we design a performance-gap reward that compares against a naive memory baseline.
Experiments on HotpotQA, 2WikiMultihopQA, and MuSiQue demonstrate that our model consistently outperforms baselines and remains robust under unseen-model replacement.
\end{abstract}

% ══════════════════════════════════════════════════════════════════════════════

\section{Introduction}
\label{sec:intro}

Recent advances in large language models have positioned LLM-based agents as a promising path toward artificial general intelligence \citep{yu2026multi,wang2024survey}.
A central challenge on this path is enabling agents to autonomously accumulate experience and evolve their capabilities over time.
Memory is the key component for addressing this challenge, and has consequently attracted significant attention from both academia and industry over the past two years \citep{bi2025stoc,feng2026elasticmem,hao2026self}.
Current memory systems are predominantly designed around a fixed LLM, tailoring operations to the specific model that consumes the memory.
Representative approaches include memory streams in generative agents \citep{park2023generative}, self-reflective verbal memory \citep{shinn2023reflexion}, skill libraries \citep{wang2023voyager}, hierarchical memory management \citep{packer2023memgpt}, and experience distillation \citep{zhao2024expel}.
These methods focus on what to store, how to compress, and when to retrieve, but all implicitly assume that the backbone remains unchanged throughout the memory lifecycle.

\begin{figure*}[t]
\centering
\includegraphics[width=\textwidth]{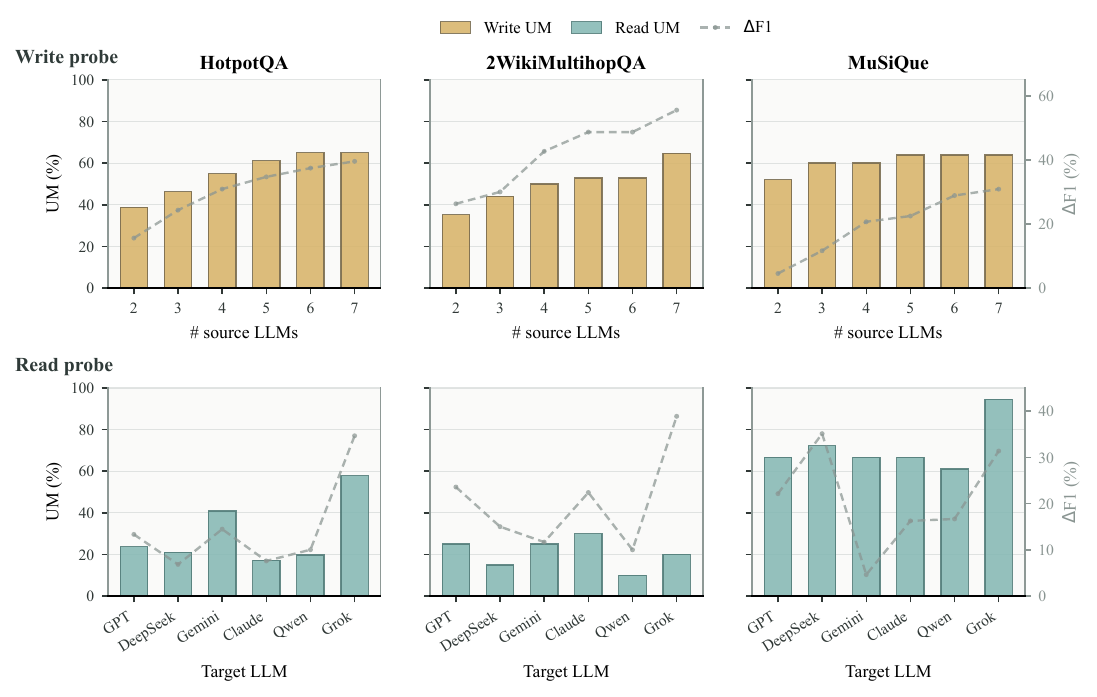}
\caption{Pilot study of memory mismatch.
Each dataset is plotted in a separate panel because absolute mismatch rates are not directly comparable across datasets or probe settings.
\textbf{Top:} write probe---the same context is processed by $K$ source LLMs; bars show unmatched answer rate (UM) as $K$ grows.
\textbf{Bottom:} read probe---the same core memory is rewritten into semantically related variants and consumed by different target LLMs.
Dashed lines show $\Delta_{\mathrm{F1}}$ (best--worst F1 gap). 
Together, the probes show that mismatch can arise both when memories are written by different source models and when similar memories are read by different target models.
The same trend appears on HotpotQA, 2WikiMultihopQA, and MuSiQue, covering two-, three-, and four-hop QA settings and suggesting that mismatch is tied to cross-model memory exchange rather than a single dataset or hop length.
Please see Appendix~\ref{sec:app:pilot} for the full protocol.}
\label{fig:pilot_mismatch}
% \vspace{-4em}
\end{figure*}

In practice, however, users often operate agents with entirely different LLMs.
Across different tasks, users may prefer Claude for coding and GPT for creative writing; even within a single task, people design routing strategies \citep{moslem2026dynamic,jitkrittum2025universal,wang2025mixture} that leverage different LLMs at different steps to exploit complementary strengths.
As a result, memory written by one LLM often needs to be consumed by another.
From the upstream LLM's write perspective, given the same preceding state, different LLMs produce different actions, leading to different content being stored in memory.
As shown in our pilot experiment (\Cref{fig:pilot_mismatch}, top row), we vary the number of source LLMs from 2 to 7 and consistently observe increasing output divergence within each dataset.
From the downstream LLM's read perspective, different memory entries, even when semantically equivalent, can cause the same LLM to produce different results, thereby affecting agent performance.
As shown in \Cref{fig:pilot_mismatch} (bottom row), semantically related memory variants also lead to different answers across target LLMs within each dataset.
These observations motivate the central question of this work: \emph{how can agent memory adapt across diverse LLMs, so that memory written by one model effectively activates the downstream LLM that consumes it?}

To answer this question, we shift the perspective from LLM-centric memory design to \emph{memory-centric LLM adaptation}, and formalize the problem of cross-LLM memory adaptation in the context of agent memory systems.
To solve this problem, we propose \textbf{Rosetta Memory (RoMem)}, which designs two LLM-aware operators from the write and read sides respectively: each operator takes an LLM profile description and the raw memory as input, and outputs adapted memory tailored to the target LLM.
We train these two operators on high-reward mixed-LLM traces.
To enable our method to support as many LLMs as possible, we construct a diverse LLM pool and design a minimum-gain sampling curriculum that prioritizes the least-served LLMs during training, progressively improving coverage.
To disentangle the contribution of our learned operators from the intrinsic capability of each LLM, we design a performance-gap reward that measures improvement over a naive memory baseline.

In summary, the main contributions of this paper are:
(1) We empirically demonstrate, from both the write and read perspectives, the necessity of memory adaptation when diverse LLMs are used in agent tasks, and formalize the problem of cross-LLM memory adaptation.
(2) We propose Rosetta Memory (RoMem), which jointly learns LLM-aware write and read operators to adapt upstream memory for downstream LLMs.
(3) We design a minimum-gain sampling curriculum and a performance-gap reward to broaden LLM coverage and isolate the operators' contribution.
(4) We conduct extensive experiments on HotpotQA, 2WikiMultihopQA, and MuSiQue, demonstrating its effectiveness.
% ══════════════════════════════════════════════════════════════════════════════
\section{Related Work}
\label{sec:related_work}

\paragraph{Memory-augmented LLM agents.}
Memory has become a central component for transforming stateless LLMs into persistent agents that can reuse observations, reflections, preferences, and interaction traces beyond the immediate prompt.
Early systems store natural-language memories for planning, dialogue, or self-reflection \citep{feng2026graphplanner,sun2026h,zhong2024memorybank,yu2026agentic}, while recent work improves memory compression, organization, lifecycle management, and online adaptation through gist memories, graph-like notes, scalable memory extraction, and episodic trajectory reuse \citep{lee2024human,chhikara2025mem0,xu2026mem,zhou2025memento,li2025memos,wang2025mirix}.
These methods address what to store, how to compress, and when to retrieve, but they generally do not explicitly treat the LLM that writes memory and the LLM that later consumes it as distinct variables, nor evaluate whether memory remains useful when the consuming LLM changes.
RoMem instead studies memory as a transferable interface between upstream writer LLMs and downstream reader LLMs within a multi-hop episode.

\paragraph{Context sensitivity and memory interfaces.}
Lightweight context learning methods, such as prefix and prompt tuning, show that compact learned context can steer frozen or partially frozen LLMs \citep{li2021prefix}, motivating our use of profile-conditioned operators around fixed black-box LLMs.
At the input level, LLM behavior is also sensitive to prompt wording, context order, and the placement of relevant evidence \citep{yan2026memory,he2024does,brucks2025prompt,feng2026elasticmem}.
This sensitivity is central to our read-side problem: even when the stored facts are unchanged, their presentation may need to adapt to the target LLM.
Closer to our work, RET-LLM introduces explicit memory read/write API calls with symbolic and vector retrieval over extracted triplets \citep{modarressi2023ret}, and MemLLM finetunes write and read models to extract relations, query memory, and condition generation on returned relations \citep{modarressi2024memllm}.
However, these methods do not separately model source and target LLM profiles, nor optimize whether memory produced by one LLM remains consumable after the reader LLM is replaced.
RoMem focuses on this cross-LLM memory adaptation problem by jointly learning profile-conditioned write and read operators.

\paragraph{Heterogeneous LLM systems.}
In practice, users and agent systems increasingly combine multiple LLMs because different models vary in cost, latency, capability, and response style.
Model routing and cascading select among candidate LLMs to balance quality and cost \citep{feng2026graphplanner,moslem2026dynamic,jitkrittum2025universal}, while ensembling and multi-agent systems aggregate outputs or organize collaboration among different models and roles \citep{hao2026self,wang2025mixture,yu2026multi,hong2024metagpt,chen2024agentverse}.
Recent work also studies structured memories for heterogeneous agents \citep{yuen2025intrinsic}.
These systems exploit heterogeneity at the level of model selection, output fusion, or role assignment.
RoMem addresses a complementary interface problem: when heterogeneous models share episode memory, the same stored text may activate one reader effectively but confuse another.
This motivates profile-conditioned memory adaptation and minimum-gain training that improves coverage for the least-served model-memory pairs rather than optimizing only average performance.

% ══════════════════════════════════════════════════════════════════════════════
\section{Preliminaries}
\label{sec:prelim}

\subsection{Memory-Involved Agent-Environment Interaction}
\label{sec:prelim:notation}

We consider an agent driven by a fixed LLM with policy $\pi$.
It interacts with an environment over $T+1$ steps indexed by $t = 0, 1, \ldots, T$.
At step $t$, the agent receives an observation $o_t \in \cO$ and produces an action $a_t$.
To support long-horizon behavior, it maintains a memory bank, denoted $M_t$ at step $t$, that evolves across the interaction.
Each step follows a read--act--write loop.
First, the agent reads a prompt-ready context $\tilde{m}_t$ from $M_t$ conditioned on $o_t$.
Then, it samples an action $a_t \sim \pi(\cdot \mid [\,o_t;\, \tilde{m}_t\,])$.
Finally, it appends a new entry summarizing $(o_t, a_t)$ to form $M_{t+1}$.
Iterating this loop over $t = 0, 1, \ldots, T$ yields a complete trajectory $\tau = (o_0, a_0, \ldots, o_T, a_T)$, which is evaluated by a terminal task reward $R(\tau)$.
The agent's objective is to maximize the expected return over trajectories, $\max\; \E_{\tau}[R(\tau)]$.

\subsection{Cross-LLM Memory Adaptation Problem}
\label{sec:prelim:problem}

As mentioned in the introduction, agents often switch backbones across steps within the same trajectory in practice.
To formally define this cross-LLM memory adaptation problem, we extend the basic memory-involved agent-environment interaction paradigm in \S\ref{sec:prelim:notation} by allowing the backbone LLM to vary across steps within the same trajectory.
At step $t$, the agent is driven by a given LLM $\ell_t$ from a heterogeneous model pool $\cL = \{L^{(1)}, \ldots, L^{(K)}\}$, and the model assignment sequence for the trajectory is denoted by $\bm{\ell} = (\ell_0, \ldots, \ell_T) \in \cL^{T+1}$.
To make memory effective across this heterogeneous pool, we parameterize the read and write operations as two learnable, LLM-aware functions: a read function $\fread(\cdot;\, \thetar)$ that takes the target LLM identifier and adapts entries from $M_t$ into a prompt-ready context, and a write function $\fwrite(\cdot;\, \thetaw)$ that takes the source LLM identifier and distills each $(o_t, a_t)$ into a memory entry.
The step-level loop becomes:
\begin{align}
    \tilde{m}_t &= \fread(o_t,\, M_t,\, \ell_t;\, \thetar), \label{eq:read}\\[2pt]
    a_t &\sim \pi_{\ell_t}\!\left(\cdot \mid [\,o_t;\; \tilde{m}_t\,]\right), \label{eq:act}\\[2pt]
    M_{t+1} &= M_t \cup \left\{\fwrite(o_t,\, a_t,\, \ell_t;\, \thetaw)\right\}. \label{eq:write}
\end{align}
Here $\ell_t$ acts as the LLM identifier that tells the operators which model will consume the read context and which model writes the current action. We specify its concrete realization in \S\ref{sec:method:operators}.

As discussed in \Cref{sec:intro}, naive memory operations become suboptimal under diverse LLMs, due to write-side divergence and read-side sensitivity.
Given an assignment $\bm{\ell}$, the read--act--write loop in \crefrange{eq:read}{eq:write} induces a trajectory distribution $P_{\bm{\ell}}^{\thetaw,\thetar}$.
The trajectory-level return is then defined as:
\begin{equation}
    J(\bm{\ell},\, \thetaw,\, \thetar) =
    \E_{\tau \sim P_{\bm{\ell}}^{\thetaw,\thetar}}
    \!\left[R(\tau)\right],
    \label{eq:return}
\end{equation}
where $R(\tau)$ is the terminal task reward on the sampled trajectory $\tau$.
Ideally, we want $\fread$ and $\fwrite$ to obtain a large $J$ under every possible assignment $\bm{\ell} \in \cL^{T+1}$.
However, directly enforcing this across all $K^{T+1}$ assignments is infeasible.
We therefore relax it into the following max--min objective:
\begin{equation}
    (\thetaw^*,\, \thetar^*) =
    \underset{\thetaw,\, \thetar}{\operatorname{arg\,max}}\;
    \min_{\bm{\ell} \in \cL^{T+1}}
    J(\bm{\ell},\, \thetaw,\, \thetar).
    \label{eq:objective}
\end{equation}
The inner $\min$ targets the worst-case assignment, so maximizing it lifts the lower bound of $J$ across all assignments.

% ══════════════════════════════════════════════════════════════════════════════
\section{Methodology}
\label{sec:method}
    
The core of our method is to learn the read and write operators $\fread$ and $\fwrite$ so that memory written by one LLM remains effective when consumed by another.
We first describe the concrete implementation of these two operators (\S\ref{sec:method:operators}), and then present their training procedure (\S\ref{sec:method:training}).

\subsection{Implementation of $\fread$ and $\fwrite$}
\label{sec:method:operators}

The write operator $\fwrite$ and the read operator $\fread$ are parameterized as two separate language models, and the LLM identifier $\ell_t$ at each step is concretely realized via a profile embedding.
Each LLM $L^{(k)} \in \cL$ is associated with a lightweight profile record containing a model identifier $i_k$, a textual description $d_k$, and metadata $\bm{q}_k$ (e.g., backend, family, scale, preferred read style).
A learnable mapping function $g$ with parameters $\theta_g$ takes this profile record to a condition vector
\begin{equation}
    \bm{e}_{L^{(k)}} = g(i_k,\, d_k,\, \bm{q}_k) \in \R^d ,
    \label{eq:profile}
\end{equation}
which is injected into $\fread$ and $\fwrite$ as soft prefix tokens \citep{li2021prefix,lester2021power}.
At step $t$, the reader uses the target-LLM embedding $\bm{e}_{\ell_t}^{\text{tgt}}$ to rewrite the bank into a prompt-ready context $\tilde{m}_t = \fread(o_t,\, M_t,\, \bm{e}_{\ell_t}^{\text{tgt}};\, \thetar)$, and the writer uses the source-LLM embedding $\bm{e}_{\ell_t}^{\text{src}}$ to distill the observation--action pair into a memory entry $m_t = \fwrite(o_t,\, a_t,\, \bm{e}_{\ell_t}^{\text{src}};\, \thetaw)$.
In essence, $g$ acts as a compression function that maps the discrete LLM identity into a continuous embedding space, which improves the model's generalization to unseen LLMs.
\subsection{Training of $\fread$ and $\fwrite$}
\label{sec:method:training}

RoMem trains $\fread$ and $\fwrite$ with an Expert Iteration-style, reward-filtered self-training loop: the current operators generate mixed-LLM rollouts, terminal rewards retain the top-$\alpha$ traces as expert demonstrations, and supervised updates imitate their read/write decisions \citep{anthony2017thinking,gulcehre2023reinforced}.
To make this loop robust across LLMs, we use two designs throughout training: a counterfactual gain that estimates each LLM's operator-specific improvement over a naive memory baseline, and a sampler that prioritizes LLMs with the lowest gain during rollout collection.
\begin{algorithm}[t]
\caption{RoMem Expert-Iteration Training with Minimum-Gain Sampling}
\label{alg:romem}
\begin{algorithmic}[1]
\REQUIRE LLMs $\cL = \{L^{(1)}, \ldots, L^{(K)}\}$, task distribution $p(\cM)$, hyperparameters $\eta$ and $\alpha$
% \STATE \textbf{Phase 0: Profile Embedding Construction}
\FOR{each $L^{(k)} \in \cL$}
    \STATE Collect profile record $(i_k,d_k,\bm{q}_k)$
    \STATE Compute profile embedding $\bm{e}_{L^{(k)}} \leftarrow g(i_k,d_k,\bm{q}_k)$ via \cref{eq:profile}
\ENDFOR
\STATE Initialize $\Delta R(L^{(k)}) \leftarrow 0$ for all $L^{(k)} \in \cL$
% \STATE \textbf{Phase 1: Supervised Read/Write Training}
\FOR{each training round until the budget is exhausted}
    \STATE Sample task $\cM \sim p(\cM)$
    \STATE Sequentially sample assignment $\bm{\ell}=(\ell_0,\ldots,\ell_T)$ using \cref{eq:schedule}
    \STATE Initialize memory bank $M_0 \leftarrow \varnothing$
    \FOR{each step $t = 0,\ldots,T$}
        \STATE $\tilde{m}_t \leftarrow \fread(o_t,M_t,\bm{e}_{\ell_t}^{\mathrm{tgt}};\thetar)$
        \STATE $a_t \sim \pi_{\ell_t}(\cdot \mid [o_t;\tilde{m}_t])$
        \STATE $m_t \leftarrow \fwrite(o_t,a_t,\bm{e}_{\ell_t}^{\mathrm{src}};\thetaw)$
        \STATE $M_{t+1} \leftarrow M_t \cup \{m_t\}$
        \STATE Log $(\ell_t,M_t,a_t,m_t,\tilde{m}_t)$
    \ENDFOR
    \STATE Rank completed trajectories by terminal reward and keep the top-$\alpha$ fraction as $\cD_{\mathrm{expert}}$
    \STATE Build $\cD_{\mathrm{train}}$ from logged write/read pairs using \cref{eq:sft_data}
    \STATE Update $\fwrite$, $\fread$, and $g$ by minimizing \cref{eq:sft}
    \STATE Periodically refresh $\Delta R(L^{(k)})$ for all $L^{(k)}$ using \cref{eq:delta_rw}
\ENDFOR
\RETURN Trained $\fwrite$, $\fread$, and profile encoder $g$
\end{algorithmic}
\end{algorithm}
\paragraph{Counterfactual gain.}
To isolate the operators' contribution from the intrinsic capability of each LLM, we define a \emph{naive baseline} memory system whose write step appends entries to the memory bank without any learned processing and whose read step concatenates all stored entries into the context.
For each LLM $L^{(k)}$, the \emph{counterfactual gain} is then
\begin{equation}
    \Delta R(L^{(k)}) = R_{f}(L^{(k)}) - R_{\text{naive}}(L^{(k)}),
    \label{eq:delta_rw}
\end{equation}
where $R_{f}(L^{(k)})$ is the expected reward over entire trajectories with $L^{(k)}$ as the backbone at every step under the current $\fread$ and $\fwrite$, and $R_{\text{naive}}(L^{(k)})$ is the analogous quantity under the naive baseline.\footnote{We measure $\Delta R$ at the trajectory level because many tasks provide only a terminal reward with no intermediate per-step signal; if per-step rewards were available, our method could be straightforwardly extended to use them.}
We refresh $\Delta R(L^{(k)})$ periodically during training using the latest checkpoints, and a small $\Delta R(L^{(k)})$ identifies an underserved LLM, regardless of its intrinsic strength.

\paragraph{Minimum-gain LLM sampling.}
To prioritize underserved LLMs during training, we sample the active model at each step according to
\begin{equation}
    p(\ell_t = L^{(k)}) \propto d_k^{(t)} \exp\!\left(-\eta\, \Delta R(L^{(k)})\right),
    \label{eq:schedule}
\end{equation}
where $d_k^{(t)} = 1$ at the first step; otherwise, letting $\cL_{\bm{\ell}_{<t}} = \{\ell_0, \ldots, \ell_{t-1}\}$ denote the LLMs already sampled within the current trajectory, we set $d_k^{(t)} = \min_{L^{(j)} \in \cL_{\bm{\ell}_{<t}}} \|\bm{e}_{L^{(k)}} - \bm{e}_{L^{(j)}}\|_2$, i.e., the embedding distance from the candidate $L^{(k)}$ to its nearest already-sampled neighbor.
The $d_k^{(t)}$ term thus promotes coverage of behaviorally diverse models within a trajectory, the exponential weight biases sampling toward those with small $\Delta R$, and $\eta > 0$ controls how strongly the sampler favors underserved LLMs.

\paragraph{Training objective.}
\label{sec:method:sft}
Consistent with the Expert Iteration loop above, we train $\fwrite$, $\fread$, and the profile encoder on high-reward mixed-LLM rollouts.
At each step, we sample $\ell_t$ via \cref{eq:schedule}, drawing both behaviorally diverse and underserved models.
During each rollout, we log the active model $\ell_t$, memory bank $M_t$, action $a_t$, written entry $m_t$, and read context $\tilde{m}_t$, rank trajectories by cumulative reward, and keep the top-$\alpha$ fraction as $\cD_{\text{expert}}$.
The training dataset aggregates the logged steps:
\begin{equation}
    \cD_{\text{train}} = \{(x_t^{w},\, m_t,\, x_t^{r},\, \tilde{m}_t) : (o_t, a_t, \ell_t) \in \tau,\; \tau \in \cD_{\text{expert}}\},
    \label{eq:sft_data}
\end{equation}
where $x_t^{w} = (o_t, a_t, \bm{e}_{\ell_t}^{\text{src}})$ and $x_t^{r} = (o_t, M_t, \bm{e}_{\ell_t}^{\text{tgt}})$.
We jointly optimize $\thetaw$, $\thetar$, and $\theta_g$ by minimizing:
\begin{equation}
    \cL_{\text{train}} = -\E_{\cD_{\text{train}}}\!\left[\log p_{\thetaw}(m_t \mid x_t^{w}) + \log p_{\thetar}(\tilde{m}_t \mid x_t^{r})\right].
    \label{eq:sft}
\end{equation}

Algorithm~\ref{alg:romem} summarizes the full RoMem training pipeline.

% ══════════════════════════════════════════════════════════════════════════════

\section{Experiments}
\label{sec:exp}
In this section, we first evaluate the overall task performance of RoMem against existing baselines, and then conduct a component-level ablation over $\fread$, $\fwrite$, and the profile embedding to identify each part's contribution.
We next study generalization to held-out LLMs and the effect of the minimum-gain LLM sampling schedule introduced in \S\ref{sec:method:training}.
Finally, we present a qualitative inspection of the read/write outputs.
We summarize these aspects as five research questions: \textbf{RQ1:} Does RoMem improve overall task performance compared with existing prompting, accumulation, and memory baselines? \textbf{RQ2:} Which RoMem components ($\fread$, $\fwrite$, and the profile embedding) contribute most to the final performance? \textbf{RQ3:} Does RoMem generalize to held-out LLMs that are unseen during training? \textbf{RQ4:} How effective is minimum-gain LLM sampling? \textbf{RQ5:} What do the learned read and write operators actually produce at the text level?

\subsection{Experimental Setup}
\label{sec:exp:setup}

\paragraph{\textbf{Datasets.}}
We take multi-hop QA as a representative testbed to validate RoMem, since it naturally involves intermediate evidence accumulation across steps and thus exposes cross-LLM memory effects clearly.
Specifically, we use HotpotQA, 2WikiMultihopQA, and MuSiQue \citep{yang2018hotpotqa,ho2020constructing,trivedi2022musique}, each adapted into fixed-length multi-hop QA trajectories.
For all the datasets, we filter out samples with fewer than the required number of hops, and use the same filtered set across all methods for a fair comparison.
After filtering and sampling, the training/test splits contain 1,851/159 examples for HotpotQA, 642/67 examples for 2WikiMultihopQA, and 481/62 examples for MuSiQue, respectively.
Following recent API-intensive multi-hop RAG evaluations that use randomly selected subsets to balance stable comparison with experimental feasibility \citep{asl2025fair}, we keep the evaluation subsets fixed across all methods.
We repeat each experiment with four random seeds and report the mean performance.

\paragraph{\textbf{Baselines.}}
% Baselines and RoMem share the same evaluation files, answer model, candidate-document source, and metric computation.
We group the baselines by how they handle memory across reasoning steps.
\textbf{(1) Memory-free baselines} access only the retrieved context without any persistent memory: Direct (answers directly from retrieved evidence), Chain-of-Thought (CoT) \citep{wei2022chain}, Scratchpad \citep{nye2021show}, Self-Ask \citep{press2023measuring}, and Least-to-Most \citep{zhou2022least}; the latter four organize retrieved evidence in-context with their respective reasoning prompts but do not persist it across steps.
\textbf{(2) Raw-buffer memory baselines} keep a running buffer of intermediate state without any learned read or write processing: Step-Accumulate concatenates step-level evidence and notes into a growing buffer, while Running-Summary \citep{lee2024human} compresses the history into a compact factual summary.
\textbf{(3) Structured memory baselines} maintain a dedicated memory store with custom organization but without backbone-aware adaptation: Mem0 \citep{chhikara2025mem0} writes per-step episodic memory entries, A-Mem \citep{xu2026mem} builds an associative network of linked notes, and Memento \citep{zhou2025memento} stores trajectory checkpoints as cases for answer generation.

\paragraph{\textbf{Implementation Details.}}
The write and read operators $\fwrite$ and $\fread$ are both initialized from the google/flan-t5-small checkpoint \citep{chung2024scaling,raffel2020exploring} for all runs.
The LLM pool of RoMem contains six API profiles: gpt-5.4-mini, gpt-5.4-nano, gemini-2.5-flash, gemini-2.5-flash-nothinking, claude-sonnet-4-6, and claude-sonnet-4-6-thinking \citep{singh2025openai,kasireddy2026evaluating, team2023gemini}.
We tune the minimum-gain sampling coefficient $\eta$ over $\{0.4,0.8,1.2,1.6,2.0\}$.
We report Exact Match (EM), token-level F1, script accuracy (Acc), and LLM-judged semantic accuracy (Acc$_{\mathrm{LLM}}$) \citep{zheng2023judging}, where Acc is a normalized containment score and Acc$_{\mathrm{LLM}}$ is a binary semantic judgment from gpt-5.4-mini.

\begin{table*}[t]
\centering
\small
\setlength{\tabcolsep}{1pt}
\caption{Main results under sample-context retrieval. Values are percentages; higher is better, and the best result in each column is shown in \textbf{bold}.}
\vspace{0.3cm}
\label{tab:main_results}
\begin{tabular*}{\textwidth}{@{\extracolsep{\fill}}lcccccccccccc@{}}
\toprule
\multirow{2}{*}{Method}
& \multicolumn{4}{c}{\textbf{HotpotQA}}
& \multicolumn{4}{c}{\textbf{2Wiki}}
& \multicolumn{4}{c}{\textbf{MuSiQue}} \\
\cmidrule(lr){2-5}\cmidrule(lr){6-9}\cmidrule(l){10-13}
& EM$\uparrow$ & F1$\uparrow$ & ACC$\uparrow$ & ACC$_{\mathrm{LLM}}\uparrow$
& EM$\uparrow$ & F1$\uparrow$ & ACC$\uparrow$ & ACC$_{\mathrm{LLM}}\uparrow$
& EM$\uparrow$ & F1$\uparrow$ & ACC$\uparrow$ & ACC$_{\mathrm{LLM}}\uparrow$ \\
\midrule
\multicolumn{13}{@{}l}{\textit{Memory-free baselines}} \\
Direct & 28.3 & 41.9 & 34.6 & 50.9 & 38.8 & 40.4 & 43.3 & 43.3 & 8.1 & 10.4 & 8.1 & 11.3 \\
CoT & 39.0 & 54.0 & 48.4 & 65.4 & 49.3 & 51.4 & 49.3 & 49.3 & 11.3 & 15.9 & 11.3 & 14.5 \\
Scratchpad & 34.6 & 48.9 & 44.0 & 56.0 & 23.9 & 28.4 & 28.4 & 28.4 & 9.7 & 15.0 & 11.3 & 17.7 \\
Self-Ask & 39.6 & 53.7 & 46.5 & 61.6 & 43.3 & 46.2 & 44.8 & 44.8 & 4.8 & 14.4 & 11.3 & 16.1 \\
Least-to-Most & 36.5 & 52.0 & 42.8 & 59.1 & 25.4 & 29.4 & 25.4 & 25.4 & 6.5 & 17.3 & 8.1 & 21.0 \\
\midrule
\multicolumn{13}{@{}l}{\textit{Raw-buffer memory baselines}} \\
Step-Accumulate & 37.7 & 53.1 & 46.5 & 64.2 & 38.8 & 41.3 & 38.8 & 38.8 & 12.9 & 21.5 & 16.1 & 21.0 \\
Running-Summary & 34.0 & 46.6 & 42.8 & 55.3 & 29.9 & 32.6 & 31.3 & 31.3 & 6.5 & 14.6 & 8.1 & 16.1 \\
\midrule
\multicolumn{13}{@{}l}{\textit{Structured memory baselines}} \\
Mem0 & 35.8 & 52.5 & 47.8 & 62.3 & 44.8 & 48.3 & 44.8 & 44.8 & 11.3 & 17.4 & 11.3 & 17.7 \\
A-Mem & 39.0 & 53.9 & 48.4 & 66.0 & 38.8 & 41.0 & 40.3 & 40.3 & 8.1 & 16.4 & 8.1 & 19.4 \\
Memento & 38.4 & 55.2 & 48.4 & 65.4 & 47.8 & 50.8 & 47.8 & 47.8 & 12.9 & 20.8 & 14.5 & 24.2 \\
\midrule
RoMem & \textbf{50.9} & \textbf{65.8} & \textbf{60.4} & \textbf{79.9} & \textbf{58.2} & \textbf{61.0} & \textbf{59.7} & \textbf{59.7} & \textbf{16.1} & \textbf{34.2} & \textbf{24.2} & \textbf{37.1} \\
\bottomrule
\end{tabular*}
\end{table*}
\subsection{RQ1: Overall Comparison}
\label{sec:exp:main}

\Cref{tab:main_results} reports the main results on the three multi-hop QA datasets.
RoMem obtains the best result in every reported column across all the datasets.
On HotpotQA, the gains are especially large under Acc$_{\mathrm{LLM}}$, suggesting that learned read/write adaptation improves semantic agreement beyond surface-form matching.
On 2WikiMultihopQA, RoMem also leads the strongest prompting and memory baselines, indicating that backbone-aware memory remains useful even when short answers make direct context-based methods competitive.
For 2WikiMultihopQA, Acc and Acc$_{\mathrm{LLM}}$ are often identical in \Cref{tab:main_results} because the answers in our filtered subset are mostly short entity strings, for which normalized containment largely agrees with semantic judging.
On MuSiQue, which is the most difficult setting in our evaluation, RoMem achieves the clearest margin on F1 and Acc$_{\mathrm{LLM}}$, showing that learned memory transformation helps preserve partial evidence across harder multi-hop chains.
Overall, the results support the central premise of RoMem: storing evidence is not sufficient by itself; the memory must also be written and read in a form that is adapted to the LLM that will consume it.
\begin{table*}[t]
\centering
\small
\setlength{\tabcolsep}{0pt}
\caption{Ablation results on the three multi-hop QA datasets. Values are percentages; higher is better, and the best result in each column is shown in \textbf{bold}.}
\vspace{0.3cm}
\label{tab:ablation_results}
\begin{tabular*}{\textwidth}{@{\extracolsep{\fill}}lcccccccccccc@{}}
\toprule
\multirow{2}{*}{Method}
& \multicolumn{4}{c}{\textbf{HotpotQA}}
& \multicolumn{4}{c}{\textbf{2Wiki}}
& \multicolumn{4}{c}{\textbf{MuSiQue}} \\
\cmidrule(lr){2-5}\cmidrule(lr){6-9}\cmidrule(l){10-13}
& EM$\uparrow$ & F1$\uparrow$ & ACC$\uparrow$ & ACC$_{\mathrm{LLM}}\uparrow$
& EM$\uparrow$ & F1$\uparrow$ & ACC$\uparrow$ & ACC$_{\mathrm{LLM}}\uparrow$
& EM$\uparrow$ & F1$\uparrow$ & ACC$\uparrow$ & ACC$_{\mathrm{LLM}}\uparrow$ \\
\midrule
Step-Accumulate
& 37.7 & 53.1 & 46.5 & 64.2
& 38.8 & 41.3 & 38.8 & 38.8
& 12.9 & 21.5 & 16.1 & 21.0 \\
\midrule
w/o $\fwrite$
& 30.2 & 45.5 & 37.7 & 52.8
& 46.3 & 47.9 & 46.3 & 46.3
& 12.9 & 25.1 & 14.5 & 29.0 \\
w/o $\fread$
& 44.7 & 60.4 & 54.1 & 74.2
& 46.3 & 49.0 & 47.8 & 47.8
& 16.1 & 27.2 & 19.4 & 30.6 \\
w/o profiles
& 30.8 & 46.0 & 35.8 & 51.6
& 44.8 & 47.0 & 44.8 & 46.3
& 11.3 & 22.6 & 14.5 & 30.6 \\
\midrule
RoMem
& \textbf{50.9} & \textbf{65.8} & \textbf{60.4} & \textbf{79.9}
& \textbf{58.2} & \textbf{61.0} & \textbf{59.7} & \textbf{59.7}
& \textbf{16.1} & \textbf{34.2} & \textbf{24.2} & \textbf{37.1} \\
\bottomrule
\end{tabular*}
\end{table*}
\subsection{RQ2: Ablation Study}
\label{sec:exp:ablation}

% {\color{red}\textbf{[Comment: Step-Accumulate looks unexpectedly strong in some columns of \Cref{tab:ablation_results}; need to double-check the protocol and re-run to confirm whether the comparison is properly controlled.]}}
We ablate the main RoMem components under the same multi-hop evaluation protocol as RQ1.
The w/o $\fwrite$ variant stores raw step evidence, w/o $\fread$ directly concatenates memory into the final context, and w/o profiles replaces source and target model profiles with a shared profile identifier.
Step-Accumulate serves as a no-read/write reference and is included as the representative non-RoMem baseline because it is consistently strong in our main comparison (\Cref{tab:main_results}); the RoMem row is copied from the main comparison.
\Cref{tab:ablation_results} shows that model profiles and the write operator are important for stable memory transfer.
Removing profiles consistently hurts performance, indicating that a shared profile is insufficient for heterogeneous backbones.
Removing $\fwrite$ also weakens most settings by leaving raw evidence less structured and less portable.
The w/o $\fread$ variant is competitive in some settings but falls behind on 2Wiki, suggesting that learned reading is most useful when raw memory needs target-specific presentation.
Together, the ablations show that profiles and learned writing are especially important, with learned reading adding robustness under harder cross-LLM transfer.
Appendix~\ref{sec:app:writer_reader_heatmap} further examines writer--reader compatibility across model pairs.

\subsection{RQ3: Transfer to Unseen Models}
\label{sec:exp:ood}

\begin{figure}[t]
\centering
\begin{subfigure}[t]{0.49\columnwidth}
\centering
\includegraphics[width=\linewidth]{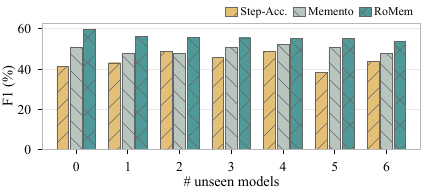}
\caption{2WikiMultihopQA}
\end{subfigure}\hfill
\begin{subfigure}[t]{0.49\columnwidth}
\centering
\includegraphics[width=\linewidth]{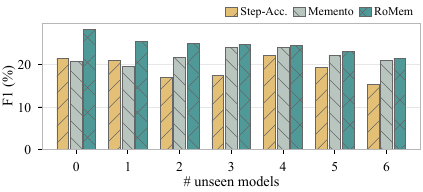}
\caption{MuSiQue}
\end{subfigure}
\caption{Model replacement on 2WikiMultihopQA and MuSiQue. Bars are grouped by the number of unseen test-time backbone models and report F1.}
\label{fig:ood_replacement}
\vspace{-1.3em}
\end{figure}

We test whether RoMem generalizes to six held-out LLMs unseen during training: \texttt{gpt-5-mini}, \texttt{deepseek-reasoner}, \texttt{gemini-2.5-flash-lite}, \texttt{grok-4-fast}, \texttt{qwen3.5-plus}, and \texttt{glm-5}.
At evaluation time, we replace $k\in\{0,\ldots,6\}$ training-pool models with held-out API models and use lightweight profiles without fine-tuning.
All methods use the same candidate-context source, answer model, and metrics; we compare RoMem with Step-Accumulate and Memento.
\Cref{fig:ood_replacement} reports F1 on 2WikiMultihopQA and MuSiQue.
On 2WikiMultihopQA, RoMem achieves the best F1 across all replacement levels, suggesting that the learned read/write interface transfers well to new LLMs.
On MuSiQue, RoMem is strongest in most settings and remains close to the best baseline otherwise, indicating that profile-conditioned adaptation does not collapse under held-out replacement.
The trends are not strictly monotonic because increasing $k$ also changes the concrete mixture of model abilities.

\subsection{RQ4: Effectiveness of Minimum-Gain Sampling}
\label{sec:exp:eta}

\begin{figure}[t]
\centering
\IfFileExists{Figs/rq4_eta_entropy.pdf}{%
\includegraphics[width=\columnwidth]{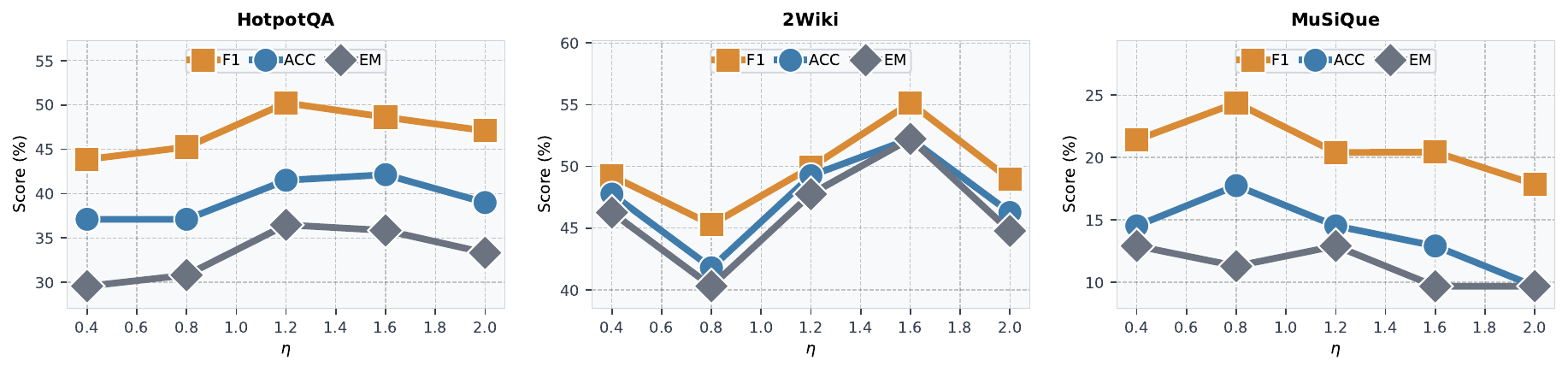}%
}{%
\fbox{\parbox{0.92\columnwidth}{\centering RQ4 eta-sensitivity figure pending rerun.}}%
}
\caption{Sensitivity to the minimum-gain sampling coefficient $\eta$ on HotpotQA, 2WikiMultihopQA, and MuSiQue. Each panel uses $\eta$ as the x-axis and reports RoMem F1, ACC, and EM.}
\label{fig:eta_sensitivity}
\vspace{-1em}
\end{figure}

This section studies the minimum-gain sampling coefficient $\eta$, which controls how strongly model sequences are biased toward backbones with smaller counterfactual gain $\Delta R$ through $\exp(-\eta\Delta R)$.
We evaluate $\eta\in\{0.4,0.8,1.2,1.6,2.0\}$ under F1, ACC, and EM on HotpotQA, 2WikiMultihopQA, and MuSiQue to inspect how different reweighting strengths affect downstream behavior.
\Cref{fig:eta_sensitivity} shows that the effect of $\eta$ is dataset-dependent rather than monotonic.
HotpotQA reaches its strongest F1 around $\eta=1.2$ and its strongest ACC around $\eta=1.6$, while 2WikiMultihopQA peaks at $\eta=1.6$ across F1, ACC, and EM.
MuSiQue favors a smaller value, with $\eta=0.8$ giving the best F1 and ACC.
These trends suggest that moderate minimum-gain reweighting is helpful, but overly strong reweighting can reduce performance by over-concentrating sampling on a narrower set of low-gain backbones.
% The remaining fluctuations indicate that textual profiles still leave room for stronger behavioral profiling.
Appendix~\ref{sec:app:worst_case_gain} reports an additional worst-case target-reader analysis.

\subsection{RQ5: Qualitative Read/Write Analysis}
\label{sec:exp:qual}

Finally, we inspect a validation example to illustrate what the write/read interface should change at the text level.
This analysis is qualitative rather than a separate benchmark: its purpose is to make the intended behavior of $\fwrite$ and $\fread$ concrete, while the quantitative contribution of these operators is measured in the ablation study in \Cref{tab:ablation_results}.
We show source memories, retrieved evidence, normalized memory, and target-conditioned read contexts derived from the same logged example.
The example in \Cref{fig:qual_examples} is selected from a 2Wiki validation case where source LLMs produce heterogeneous intermediate memories, making it useful for inspecting whether a transferable memory should preserve evidence while avoiding direct answer leakage.
Appendix~\ref{sec:app:noisy_memory} further tests whether the reader can handle polluted memory banks.

\begin{figure}[t]
\centering
\romemqualbox{Logged source memory plus evidence $\rightarrow$ normalized written memory}{
\textbf{Dataset.} 2WikiMultihopQA.\\
\textbf{Question.} Which film whose director is younger, \emph{Hell to Pay} or \emph{Darwin's Nightmare}?\\
\textbf{Verbatim source memory.} GPT-5.4-mini: ``Need compare directors' ages: Jay Jennings was born in 1965, and Hubert Sauper's birth year is missing---look up Hubert Sauper's birth date/year to determine which director is younger.''\\
\textbf{Verbatim retrieved evidence.} ``Jay Robert Jennings( born August 23, 1965)''; ``Hubert Sauper( born 27 July 1966)''.\\
\textbf{Normalized written memory.} \emph{Hell to Pay}: director Jay Jennings, born 1965. \emph{Darwin's Nightmare}: director Hubert Sauper, born 1966. Need compare which director is younger.\\[2pt]
\textcolor{romemSlate}{\emph{Effect:} removes the search instruction and most source-model uncertainty, keeps the two comparable facts, and avoids writing the final film answer.}
}
\vspace{0.55em}
\romemqualbox{Same memory $\rightarrow$ profile-conditioned read contexts}{
\textbf{Target: \texttt{gpt-5.4-nano}.}
\emph{Hell to Pay} -- Jay Jennings -- 1965. \emph{Darwin's Nightmare} -- Hubert Sauper -- 1966. Pick the younger director's film; answer as a title.\\
\textbf{Target: \texttt{claude-sonnet-4-6-thinking}.}
Use the memory as an evidence chain: map each film to its director, compare 1965 vs. 1966, then decide which film has the younger director. Do not rely on outside facts.\\
\textbf{Target: \texttt{gemini-2.5-flash-nothinking}.}
Table: \emph{Hell to Pay} | Jay Jennings | 1965. \emph{Darwin's Nightmare} | Hubert Sauper | 1966. Short answer only.\\[2pt]
\textcolor{romemSlate}{\emph{Effect:} keeps the same facts but varies ordering, explicitness, and answer-format constraints according to the target profile.}
}
\caption{Representative read/write inspection from a held-out 2Wiki validation sample. 
% The example shows how source memory and retrieved evidence can be normalized and then adapted for different target profiles.
}
\label{fig:qual_examples}
\vspace{-1em}
\end{figure}

\paragraph{\textbf{Failure modes.}}
We also observe several common failure cases.
The writer can inherit incomplete source memories, preserve a source model's uncertainty, or copy a final-answer statement into memory.
The reader can over-compress the stored evidence and remove a bridge fact needed by weaker target models.
These cases suggest that $\fwrite$ should favor neutral evidence over conclusions, while $\fread$ should adapt presentation without discarding task-critical facts.

\section{Conclusion}
\label{sec:conclusion}

This paper studies cross-LLM memory adaptation, where memory written under one backbone must be consumed by another.
We propose Rosetta Memory (RoMem), a profile-conditioned interface that learns to write transferable evidence and read it in a target-aware form.
Across three multi-hop QA datasets, RoMem improves over prompting, accumulation, and memory baselines, supporting a memory-centric view of heterogeneous LLM agents.
Future work can extend this interface to longer-horizon tasks and richer behavioral profiles.

% \newpage
\bibliography{reference}
\bibliographystyle{iclr2026_conference}
\newpage
\appendix
\section{Limitations}
\label{sec:app:limitations}

RoMem is a first step toward memory-centric adaptation for heterogeneous LLM agents, and our study has several limitations.

\begin{itemize}
    \item We evaluate controlled multi-hop QA episodes with fixed evidence-chain lengths. This setting exposes cross-LLM memory effects clearly, but it does not fully cover open-ended web navigation, tool use, dialogue, or very long-horizon planning.
    \item Our experiments use filtered evaluation subsets and a finite pool of seen and held-out API LLMs because each example requires many writer, reader, baseline, and judge calls. Larger model pools and larger evaluation sets would further strengthen the empirical picture.
    \item The learned reader and writer can still fail by over-compressing memory, preserving source-model uncertainty, or producing weak target-specific contexts.
\end{itemize}

\section{Broader Impact}
\label{sec:app:broader_impact}

RoMem aims to make heterogeneous LLM systems more reliable when intermediate evidence is shared across models.
Better cross-LLM memory transfer could reduce unnecessary recomputation, make lower-cost models more useful inside multi-step workflows, and improve robustness when systems switch models because of latency, availability, or cost constraints.
The write/read interface may also improve inspectability by converting source-specific traces into compact evidence and by exposing the context presented to each target LLM.

The same capability also creates risks.
If memory contains incorrect, biased, private, or copyrighted content, a learned writer may preserve it and a learned reader may rephrase it in a more authoritative form.
Cross-model transfer can further spread such errors across systems that otherwise would not share the same internal state.
Practical deployments should therefore use data minimization, access control, deletion mechanisms, audit logs, and human inspection for persistent memories.
For high-stakes domains, automatically rewritten memory should be treated as supporting evidence rather than as a trusted source of truth, and final decisions should rely on independently verifiable information.

\section{Experimental Setting Details}
\label{sec:app:exp_details}

\subsection{Dataset Construction and Episode Format}
We convert each QA instance into an episode with a fixed number of reasoning hops.
HotpotQA, 2WikiMultihopQA, and MuSiQue \citep{yang2018hotpotqa,ho2020constructing,trivedi2022musique} are instantiated as 2-hop, 3-hop, and 4-hop episodes, respectively, matching their typical evidence-chain lengths in our processed files.
Examples with fewer usable hops than the target episode length are filtered out before training and evaluation.
After filtering and sampling, the training/test splits contain 1,851/159 examples for HotpotQA, 642/67 examples for 2WikiMultihopQA, and 481/62 examples for MuSiQue, respectively.
All methods use the candidate context attached to the current example rather than an external Wikipedia dump, and gold supporting sentences are not directly inserted as prompts during evaluation.
At hop $t$, the model receives the question, the current candidate evidence, and the accumulated state produced by the method being evaluated.
The final answer is generated only after the last hop, so all methods share the same episode length and answer interface.

\subsection{Backbone LLM Pool and Profiles}
The seen backbone pool contains six API models: \textbf{gpt-5.4-mini}, \textbf{gpt-5.4-nano}, \textbf{gemini-2.5-flash}, \textbf{gemini-2.5-flash-nothinking}, \textbf{claude-sonnet-4-6}, and \textbf{claude-sonnet-4-6-thinking} \citep{singh2025openai,kasireddy2026evaluating,team2023gemini}.
The held-out pool used in the model-replacement experiment contains six unseen API models: \textbf{gpt-5-mini}, \textbf{deepseek-reasoner}, \textbf{gemini-2.5-flash-lite}, \textbf{grok-4-fast}, \textbf{qwen3.5-plus}, and \textbf{glm-5}.
When $k$ unseen models are evaluated, the first $k$ models in this fixed held-out list replace the corresponding seen models.
Each model is represented by a lightweight profile record consisting of its model name, a short textual description, and metadata such as backend, model family, scale bucket, and preferred read style.
For held-out models in the model-replacement experiment, we construct the same type of profile record and use it directly at test time.
No task rollouts or gradient updates are performed for the held-out models.

\subsection{Training Details}
RoMem is trained with Expert Iteration-style supervised fine-tuning from mixed-model rollouts.
Each iteration collects read/write traces, filters high-reward trajectories, and uses minimum-gain scheduling to sample more often from models whose current read--write gain is low.
We run two outer Expert-Iteration rounds.
Each round samples 10 mixed-model rollouts per training example, keeps the top-$\alpha$ fraction of reward-ranked trajectories with $\alpha=0.3$ and a minimum of one retained trajectory per example, and then trains the write/read operators for two SFT epochs.
The rollout assignments use the minimum-gain sampler with $\eta=1.2$ unless otherwise specified.

The profile encoder $g$ maps each profile to a 256-dimensional condition vector.
It combines three inputs: a learned model-ID embedding, a mean-pooled embedding of the textual profile description, and an MLP over profile metadata.
The metadata consists of the backend type, model family, scale bucket, preferred read style, a log-scale feature, a scale-present flag, a normalized model-name hash, and normalized description length.
The metadata MLP uses hidden size 128; the description token embedding size is 128 with a maximum description length of 48 tokens.
The concatenated representation is passed through a two-layer fusion MLP with GELU activations and layer normalization.
For both $\fwrite$ and $\fread$, the resulting profile vector is projected into eight soft prefix tokens and prepended to the encoder input embeddings of the corresponding \texttt{google/flan-t5-small} sequence-to-sequence model.
The writer and reader use separate FLAN-T5-small backbones and separate prefix projections.

We optimize the trainable read/write and profile-conditioned components with AdamW using batch size 4, gradient accumulation 4, weight decay 0.01, warmup ratio 0.03, maximum source length 512, and maximum target length 160.
We sweep learning rates in $\{1{\times}10^{-5},2{\times}10^{-5},3{\times}10^{-5}\}$ and tune the minimum-gain sampling coefficient $\eta$ over $\{0.4,0.8,1.2,1.6,2.0\}$.
All experiments are run on NVIDIA A800-SXM4-80GB GPUs.

\subsection{Baseline Descriptions}

\paragraph{Prompting and decomposition baselines.}
\begin{itemize}
    \item \textbf{Direct} answers the question directly from the shared candidate context without explicit decomposition or memory.
    \item \textbf{CoT} \citep{wei2022chain} prompts the model to produce intermediate reasoning before answering.
    We instantiate it with the same candidate context and final-answer interface as the other context-using baselines.
    \item \textbf{Scratchpad} \citep{nye2021show} externalizes intermediate computation as textual notes.
    We use it as a multi-step scratch space over the retrieved candidate evidence.
    \item \textbf{Self-Ask} \citep{press2023measuring} decomposes a complex question into follow-up subquestions and answers them before producing the final answer.
    We implement the follow-up process over the per-example candidate context rather than external search.
    \item \textbf{Least-to-Most} \citep{zhou2022least} solves simpler subproblems first and uses their answers to solve the original question.
    We instantiate the decomposition within the fixed hop budget of each dataset.
\end{itemize}

\paragraph{State accumulation baselines.}
\begin{itemize}
    \item \textbf{Step-Accumulate} keeps the raw evidence and notes collected at previous hops.
    In our setting, it appends hop-level evidence without learned write or read operators.
    \item \textbf{Running-Summary} \citep{lee2024human} maintains a compact summary of the reasoning history.
    We update this summary across hops and pass the final summary to the answer model.
\end{itemize}

\paragraph{Memory baselines.}
\begin{itemize}
    \item \textbf{Mem0} \citep{chhikara2025mem0} treats useful past information as retrievable memories for later use.
    We adapt it to the episode-local QA setting by storing hop evidence as memories within the current question.
    \item \textbf{A-Mem} \citep{xu2026mem} organizes agent experience as structured notes that can be retrieved in later steps.
    We adapt each hop into a note-style memory entry.
    \item \textbf{Memento} \citep{zhou2025memento} represents experience as an episodic trajectory that can guide later decisions.
    We instantiate it by treating the hop sequence of the current QA instance as the trajectory memory.
\end{itemize}

\subsection{Evaluation Protocol}
All methods are evaluated on the same dataset subsets, candidate-context source, answer model, and metric implementation.
We report EM, token-level F1 \citep{rajpurkar2016squad,yang2018hotpotqa}, Acc, and Acc$_{\mathrm{LLM}}$ \citep{zheng2023judging}.
EM and F1 follow standard QA normalization.
Acc is a normalized containment score that checks whether the gold answer appears in the normalized prediction.
Acc$_{\mathrm{LLM}}$ is a binary semantic judgment from gpt-5.4-mini that compares the question, gold answer, and prediction.

\section{Pilot Study Details}
\label{sec:app:pilot}

We run the controlled mismatch study on the same three multi-hop QA datasets used in the main experiments.
Gold hop sentences are used only in this diagnostic study to construct controlled intermediate states and canonical memories.
They are not inserted into the prompts of the main evaluation.

\paragraph{Write-side probe.}
The goal is to test whether different source LLMs write interchangeable memory when they receive the same multi-hop state.
For each example, we build a penultimate state from the question and all supporting hops except the last one.
Each source LLM is then asked to produce the next compact action or memory entry for the missing bridge.
The source pool is ordered as \textbf{gpt-5.4-mini}, \textbf{deepseek-reasoner}, \textbf{gemini-2.5-flash}, \textbf{claude-sonnet-4-6}, \textbf{qwen3.5-plus}, \textbf{grok-4-fast}, and \textbf{gpt-5-mini}.
We vary the active source pool size from 2 to 7 by taking the first $k$ models in this fixed order.
Each written memory is paired with the same known evidence and passed to a fixed answer model, \textbf{gpt-5.4-mini}, so answer variation mainly reflects differences in the source-written memory rather than differences in the final answerer.

\paragraph{Read-side probe.}
The goal is to test whether a target LLM is sensitive to semantically equivalent memory presentations.
For each example, we first construct a canonical memory from the question and the full set of supporting hops.
A rewrite model, \textbf{gpt-5.4-mini}, generates exactly four alternative memory rewrites that should preserve the same entities, numbers, dates, and relations.
We keep only rewrites that pass local semantic-preservation filters based on key-term coverage and token-overlap with the canonical memory.
The default thresholds are 0.25 for key-term coverage and 0.20 for token-overlap, with at most three rewrite attempts.
Each accepted rewrite is then fed separately to six target LLMs: \textbf{gpt-5.4-mini}, \textbf{deepseek-reasoner}, \textbf{gemini-2.5-flash}, \textbf{claude-sonnet-4-6}, \textbf{qwen3.5-plus}, and \textbf{grok-4-fast}.
Unlike the write-side probe, there is no separate fixed answer model here; the target LLM both reads the memory variant and produces the final answer.

\paragraph{Metrics and aggregation.}
We report unmatched rate (UM) and $\Delta_{\mathrm{F1}}$.
UM is the fraction of examples for which at least two variants produce non-matching final answers.
The matching rule is local rather than LLM-judged: it normalizes punctuation, articles, and whitespace, checks dataset answer aliases, treats matching numeric answers as equivalent, and also accepts high token-overlap matches.
$\Delta_{\mathrm{F1}}$ is the difference between the best and worst token-level F1 among variants for the same example, computed against the gold answer and its aliases.
The reported pilot curves average these quantities over the selected examples, with at most two repeated runs under different random seeds for robustness.

\paragraph{Results.}
\Cref{fig:pilot_mismatch} shows that memory mismatch appears from both the write and read sides.
In the write-side probe, increasing the number of source LLMs from 2 to 7 raises UM from 38.8 to 65.0 on HotpotQA, from 35.3 to 64.7 on 2WikiMultihopQA, and from 52.0 to 64.0 on MuSiQue.
The corresponding $\Delta_{\mathrm{F1}}$ also increases, reaching 39.6, 55.6, and 30.9 points on the three datasets, respectively.
In the read-side probe, semantically related memory rewrites still lead to substantial answer variation across target LLMs, with average UM of 30.0 on HotpotQA, 20.8 on 2WikiMultihopQA, and 71.3 on MuSiQue.
These trends support the motivation that both writing memory and reading memory should be conditioned on the LLM that produces or consumes it.

\section{Additional Experimental Results}
\label{sec:app:additional_results}

\subsection{Writer--Reader Compatibility Heatmap}
\label{sec:app:writer_reader_heatmap}

\paragraph{Experimental Setup.}
This diagnostic experiment measures whether memory written by one LLM backbone remains easy to consume by another backbone.
We run the experiment on HotpotQA and 2WikiMultihopQA.
To make the writer--reader mismatch visible, we use seven heterogeneous API backbones: \textbf{gpt-5.4-mini}, \textbf{claude-sonnet-4-6}, \textbf{gemini-2.5-flash}, \textbf{deepseek-reasoner}, \textbf{qwen3.5-plus}, \textbf{gpt-4.1-mini}, and \textbf{gpt-4o-mini}.
For each dataset, we sample 34 evaluation examples.
For each example and each ordered writer--reader pair, the writer receives the question together with retrieved context paragraphs rather than oracle gold evidence, and is instructed to store only intermediate evidence without stating the final answer or final choice.
The reader then consumes that memory and generates the final answer.
This gives a $7\times7$ compatibility matrix, where rows correspond to memory writers and columns correspond to memory readers.
We evaluate two memory conditions.
In the raw-memory condition, the reader directly consumes the writer's unmodified memory.
In the RoMem condition, the same raw memory is first canonicalized by the SFT-trained $\fwrite$ using the writer profile and then adapted by the SFT-trained $\fread$ using the reader profile.
Each heatmap cell reports the average token-level F1, so the visualization directly shows both absolute answer quality and writer--reader compatibility.
We also compare the diagonal and off-diagonal averages: a strong diagonal indicates that memories are easiest for the same backbone to read, while a smaller diagonal--off-diagonal gap indicates more uniform cross-backbone consumption.

\begin{figure*}[t]
\centering
\IfFileExists{Figs/draft_writer_reader_diagonal_heatmap.pdf}{%
\includegraphics[width=\textwidth]{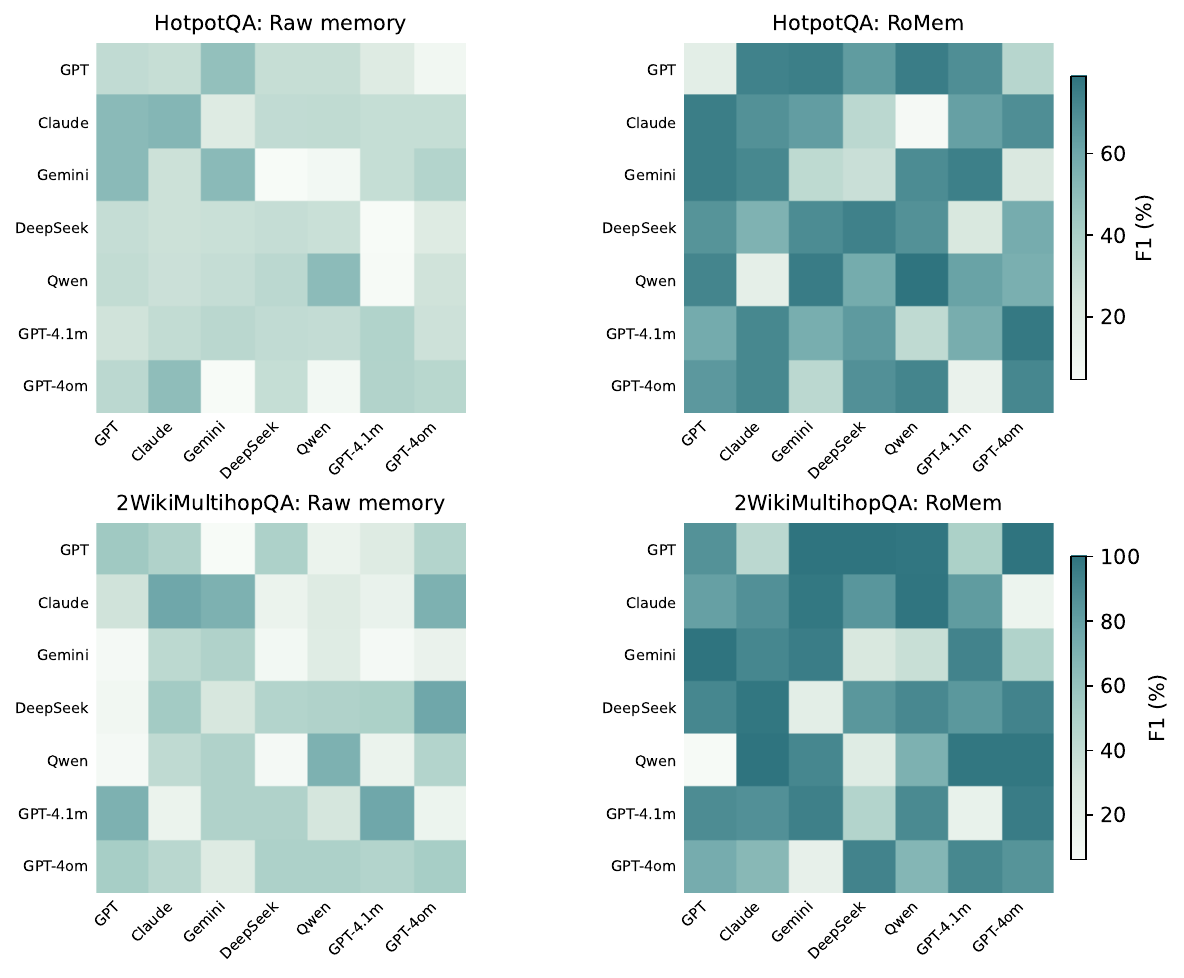}%
}{%
\fbox{\parbox{0.92\textwidth}{\centering Writer--reader compatibility heatmap pending.}}%
}
\caption{Writer--reader compatibility heatmaps on HotpotQA and 2WikiMultihopQA. Each cell reports average token-level F1 under retrieved-context memory writing. Rows denote source memory writers and columns denote target readers.}
\label{fig:app_writer_reader_heatmap}
\vspace{-0.6em}
\end{figure*}

\paragraph{Results.}
\Cref{fig:app_writer_reader_heatmap} visualizes whether a memory written by one backbone can be reliably consumed by another.
Under raw memory, the heatmaps show a clear same-backbone preference: the diagonal/off-diagonal F1 gaps are 13.7 points on HotpotQA and 25.9 points on 2WikiMultihopQA.
This indicates that raw memory is much easier to reuse when the writer and reader are the same or closely aligned.
RoMem produces a more even compatibility matrix while also improving average F1, from 29.7 to 57.3 on HotpotQA and from 38.8 to 75.1 on 2WikiMultihopQA.
The corresponding RoMem gaps are close to zero ($-0.4$ and $-0.2$ points), suggesting that the learned write/read interface substantially reduces the writer-specific diagonal pattern.

\subsection{Worst-Case Gain Analysis}
\label{sec:app:worst_case_gain}

\paragraph{Experimental Setup.}
This diagnostic experiment complements the average scores in \Cref{tab:main_results} by measuring whether RoMem improves the worst served target readers under the minimum-gain motivation in \S\ref{sec:method:training}.
We run the analysis on HotpotQA and 2WikiMultihopQA with 34 held-out evaluation examples per dataset.
The model pool is the same six seen API profiles used in the main experiments: \textbf{gpt-5.4-mini}, \textbf{gpt-5.4-nano}, \textbf{gemini-2.5-flash}, \textbf{gemini-2.5-flash-nothinking}, \textbf{claude-sonnet-4-6}, and \textbf{claude-sonnet-4-6-thinking}.
For each target reader $t$, we fix the final model in the assignment sequence to $t$ and randomly sample three source-to-target assignment sequences from the remaining seen profiles using a fixed seed.
HotpotQA uses two-step assignments of the form source $\rightarrow$ target, while 2WikiMultihopQA uses three-step assignments of the form source$_1 \rightarrow$ source$_2 \rightarrow$ target, with no repeated source model within a sequence.
For each sequence and each example, we evaluate Step-Accumulate and RoMem under the same assignment.
Step-Accumulate passes the accumulated raw step notes to the final target reader, while RoMem stores each source step through $\fwrite$ and adapts the memory bank to the final target through $\fread$.
For each target reader, we compute token-level F1 for each sampled assignment and report the worst F1 obtained by Step-Accumulate and RoMem across the three assignments.

\paragraph{Results.}
\begin{figure}[t]
\centering
\includegraphics[width=\textwidth]{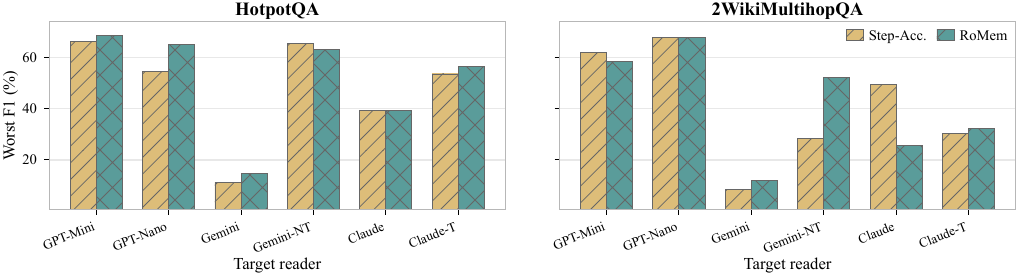}
\caption{Worst-case target-reader performance on HotpotQA and 2WikiMultihopQA. For each target reader, we report the lowest F1 across three sampled source-to-target assignment sequences.}
\label{fig:app_worst_case_gain}
\end{figure}

\Cref{fig:app_worst_case_gain} shows that RoMem improves the worst-sequence F1 for four of six target readers on HotpotQA and three of six target readers on 2WikiMultihopQA.
On HotpotQA, RoMem raises the worst-case score for GPT-Mini, GPT-Nano, Gemini, and Claude-Thinking, with the largest gain appearing for GPT-Nano.
On 2WikiMultihopQA, the largest improvement appears for Gemini-NT, where RoMem substantially raises the weakest assignment, followed by smaller gains for Gemini and Claude-Thinking.
The result also reveals residual failure cases: GPT-Mini and Claude on 2Wiki, together with a few nearly tied targets, still favor Step-Accumulate.
Thus, RoMem improves many worst-served readers, but the learned interface does not eliminate all assignment-level regressions.

\subsection{Noisy Memory Robustness}
\label{sec:app:noisy_memory}

\paragraph{Experimental Setup.}
This diagnostic experiment tests whether RoMem's read operator can filter useful evidence from a polluted memory bank, a common failure mode in long-running agent memory.
Following the motivation of noisy-memory evaluations in agent memory studies, we inject distractor, stale, and conflicting facts into the memory bank and compare raw memory consumption against RoMem's profile-conditioned reading.
We run the experiment on HotpotQA and 2WikiMultihopQA with 34 held-out examples per dataset.
For the target readers, we use three representative seen API profiles from the main model pool: \textbf{gpt-5.4-mini}, \textbf{gemini-2.5-flash}, and \textbf{claude-sonnet-4-6}.
For each example, the clean memory bank is initialized from the gold hop evidence.
We then inject $n\in\{0,1,2,3\}$ noisy memory entries and shuffle the resulting memory order.
The injected entries cycle through three noise types: distractor facts sampled from non-supporting retrieved context paragraphs, stale facts produced by perturbing dates, numbers, or entity attributes in clean evidence, and conflicting facts that contradict a key relation or answer-bearing statement.
Both methods receive the same shuffled noisy memory bank.
The raw-memory baseline directly concatenates the memory bank and passes it to the target reader, while RoMem first applies $\fread$ with the target profile and then passes the adapted context to the same target reader.
For each dataset, target reader, and noise level, we report the average token-level F1 over the 34 examples.

\paragraph{Results.}
\begin{figure}[t]
\centering
\includegraphics[width=\textwidth]{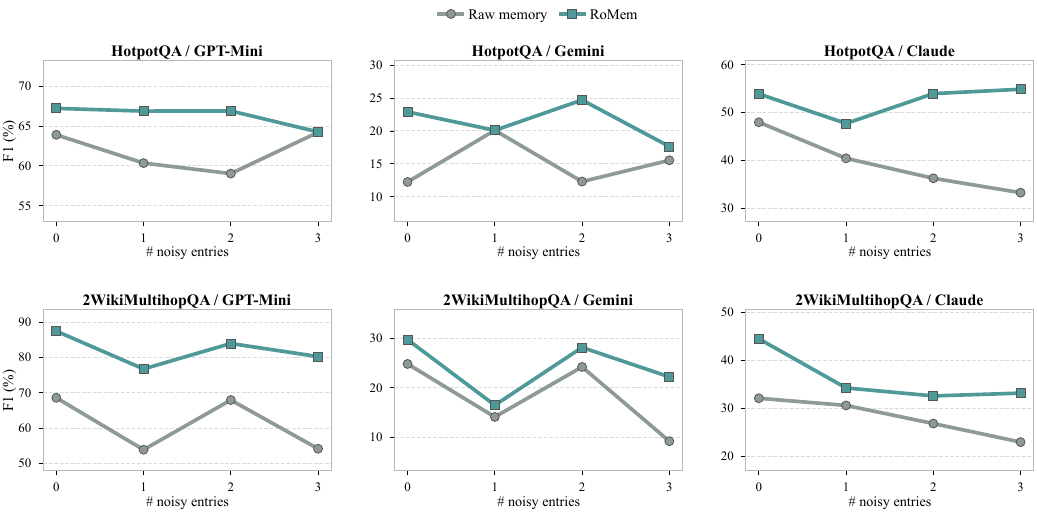}
\caption{Noisy memory robustness on HotpotQA and 2WikiMultihopQA across three target readers. The x-axis is the number of injected noisy memory entries, and the y-axis reports F1.}
\label{fig:app_noisy_memory}
\end{figure}

\Cref{fig:app_noisy_memory} shows that RoMem achieves higher F1 than raw memory for nearly all target-reader and noise-level combinations.
On HotpotQA, average gains over the four noise settings are 4.4, 6.3, and 13.1 F1 points for GPT-Mini, Gemini, and Claude, respectively.
On 2WikiMultihopQA, the gains are 21.0, 6.1, and 8.0 points, again with the clearest and most stable margin on GPT-Mini.
The RoMem curves generally stay above raw memory as more distractor, stale, or conflicting entries are added, although they are not strictly monotonic with the amount of injected noise.
These trends support the role of $\fread$ as a target-conditioned filter that can reduce the impact of distractor, stale, and conflicting memory entries.

% ══════════════════════════════════════════════════════════════════════════════
\end{document}